\documentclass{article} 
\usepackage{nips13submit_e,times}
\usepackage{hyperref}
\usepackage{url}
\usepackage{graphicx}
\usepackage{subfigure}

\title{Detecting Faces Using Region-based Fully Convolutional Networks}

\author{
  Yitong Wang \quad Xing Ji \quad Zheng Zhou \quad Hao Wang \quad Zhifeng Li\thanks{Corresponding author}\\
  Tencent AI Lab, China\\
  \texttt{\{yitongwang,denisji,encorezhou,hawelwang,michaelzfli\}@tencent.com} \\
}

\newcommand{\ignore}[1]{}

\nipsfinalcopy 

\begin{document}

\maketitle

\begin{abstract}

Face detection has achieved great success using the region-based methods.
In this report, we propose a region-based face detector applying deep networks in a fully convolutional fashion, named Face R-FCN. Based on Region-based Fully Convolutional Networks (R-FCN), our face detector is more accurate and computationally efficient compared with the previous R-CNN based face detectors. In our approach, we adopt the fully convolutional Residual Network (ResNet) as the backbone network. Particularly, we exploit several new techniques including position-sensitive average pooling, multi-scale training and testing and on-line hard example mining strategy to improve the detection accuracy. Over two most popular and challenging face detection benchmarks, FDDB and WIDER FACE, Face R-FCN achieves superior performance over state-of-the-arts.

\end{abstract}

\section{Introduction}
\begin{figure}
  \centering
  \includegraphics[width=13cm, keepaspectratio]{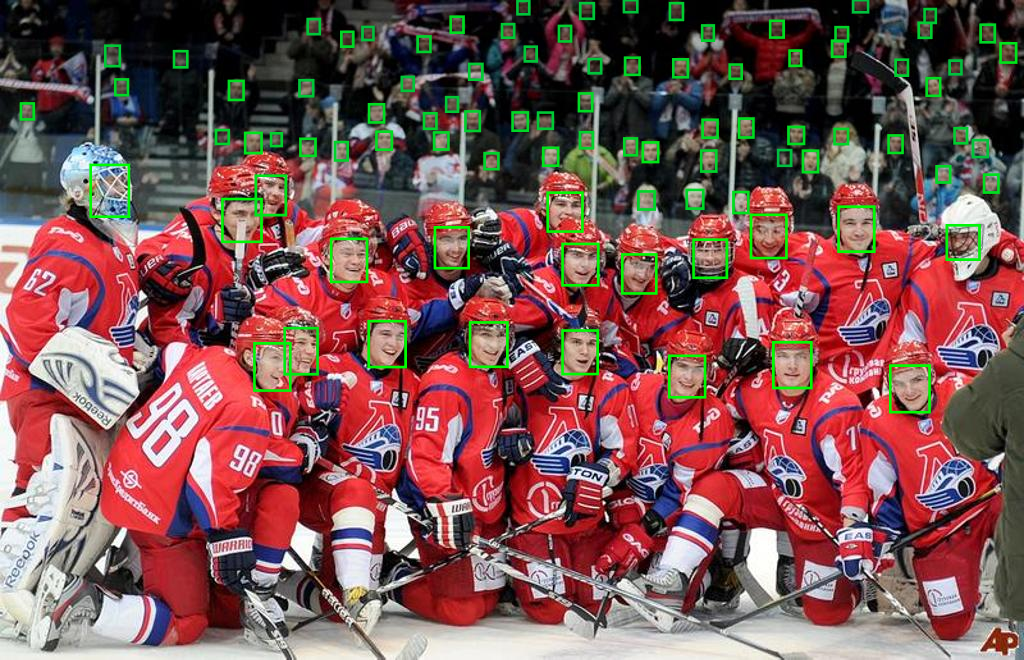}
  \caption{An example image which has extreme variability in the face regions. Green frames stand for the detection results of the proposed face detector.}\label{1}
\end{figure}

Face detection plays an important role in the modern face-relevant applications. Despite the great progress made in recent years, the technical challenging of face detection still exists out of the complex variations of real-world face images.
As shown in Figure \ref{1}, the visual faces vary a lot as the result of the affecting factors including occlusion on the facial part, different scales, illumination conditions, various poses of person, rich expressions, etc.
Recently, remarkable advances of objection detection have been driven by the success of region-based methods \cite{RCNNs,RCNN,FastRCNN,FasterRCNN}. Among recent novel algorithms, Fast/Faster R-CNN \cite{FastRCNN,FasterRCNN} are representative R-CNN based methods that perform region-wise detections on the regions of interest (RoIs). However, directly applying the strategy of region-specific operation to fully convolutional networks, such as Residual Nets (ResNets) \cite{ResNet101}, results in inferior detection performance owing to the overwhelming classification accuracy. In contrast, R-FCN \cite{RFCN} is proposed to address the problem in the fully convolutional manner.
The ConvNet of R-FCN is built with the computations shared on the entire image, which leads to the improvement of training and testing efficiency. Comparing with R-CNN based methods, R-FCN proposes much fewer region-wise layers to balance the learning of classification and detection for naturally combining fully convolutional network with region-based module.

As a specific area of generic object detection, face detection has achieved superior\ignore{high} performance thanks to\ignore{due to} the appearance\ignore{applying } of region-based methods. Previous works primarily\ignore{mainly} focus on the R-CNN based methods and achieve\ignore{making for} promising results. In this report, we develop a face detector on the top of R-FCN with elaborate design of the details, which achieves more decent performance than the R-CNN face detectors \cite{facefrcnn, facercnn}. According to the size of the general face, we carefully design size of anchors and RoIs. Since \ignore{that} the contribution of facial parts may be different for detection, we introduce a position-sensitive average pooling to generate embedding features for enhancing discrimination \ignore{to inject embedding in the feature}, and eliminate\ignore{eliminating} the effect of non-uniformed contribution in each facial part\ignore{part}.
Furthermore, we also apply\ignore{employ and} the multi-scale training and testing strategy in this work. The on-line hard example mining (OHEM) technique \cite{OHEM} is\ignore{also adopted} integrated into our network as well for boosting the learning on hard examples.\ignore{improving performance}

Our key contributions are summarized below:

(1) We develop a face detection framework that takes the special properties of face into account by integrating several innovative and effective \ignore{newly developed} techniques. The proposed approach is based on R-FCN and is well suited for face detection, thus we call it Face R-FCN.

(2) We introduce a novel position-sensitive average pooling to re-weight embedding responses on score maps and eliminate the effect of non-uniformed contribution in each facial part.

(3) By far, the proposed algorithm is benchmarked on WIDER FACE dataset \cite{wider} and FDDB dataset \cite{fddb}. Our Face R-FCN has reached the first-rate performance over the state-of-the-arts on both datasets.


\section{Related Work}


In the past decades, face detection has been extensively studied.
The pioneering work of Viola and Jones \cite{vj} 
invents a cascaded AdaBoost face detector using Haar-like features.
\ignore{proposed to learn the cascaded AdaBoost classifier based on Haar-like features for face detection.}
After that, numerous of works \cite{vj1,vj2,vj3} have focused on developing more advanced features and more powerful classifiers. 
Besides the boosted cascade methods, several studies apply deformable
part models (DPM) for face detection \cite{dpmface2,dpmface3,dpmface}. 
\ignore{Besides the cascade based face detection methods, \cite{dpmface2,dpmface3,dpmface} use the deformable
part models (DPM) for face detection.} 
The DPM methods detect faces by modeling the relationship of deformable facial parts.
\ignore{define a face
with a collection of deformable parts and train a classifier to find these parts and their relationship.}

Recent progress in face detection mainly benefits from the powerful 
deep learning approaches. The CNN-based detectors have achieved the highest
performance. \cite{cascadeCNN,cascadeCNN2,spl} construct cascaded CNNs to learn face detectors with a coarse-to-fine
strategy.  MTCNN \cite{spl} develops a multi-task training framework to jointly learn the face detection and alignment.
UnitBox \cite{unitbox} propose the intersection-over-union (IoU) loss function, to directly minimize the
IoUs of the predictions and the ground-truths. Recently, 
several methods \cite{facefrcnn,xm,deepir,cmsrcnn} use the Faster R-CNN framework to improve the face detection performance.
\cite{HR} explores the contextual information for face detection and proposes a framework achieving high performance, especially improving the accuracy of tiny faces. Most recently, \cite{ssh,sfd} propose to use single stage framework for face detection, with carefully designed strategies and achieve the state-of-the-art performance.

Similar to face detection, general object detection is advancing rapidly thanks to the deep learning approaches. 
Typical\ignore{Representative} work including R-CNN \cite{RCNNs}, Fast R-CNN \cite{FastRCNN}, Faster R-CNN \cite{FasterRCNN} and their extensions \cite{RFCN,FPN,maskrcnn}. Among these studies, R-FCN makes the detection in a nearly fully convolutional manner, which greatly enhances the efficiency of training and testing.
The methods of hard example mining further help deep learning based object detection to improve the performance. In \cite{OHEM}, the authors proposed an on-line hard example mining (OHEM) algorithm to improve the object detection performance. \cite{xm,deepir,spl} also use hard example mining algorithms to boost the performance of face detection.

\section{Proposed Approach}

\begin{figure*}
  \centering
  \includegraphics[width=13cm, keepaspectratio]{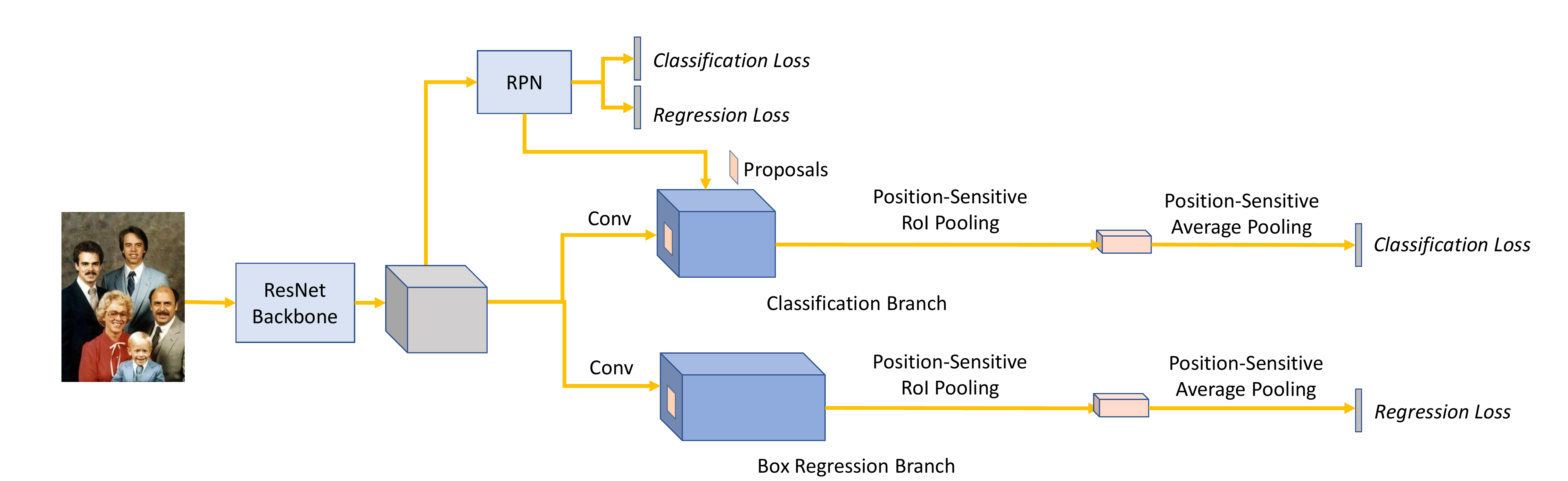}
  \caption{An overview of our R-FCN based framework. Note that position-sensitive average pooling is used to replace global average pooling for the final feature voting.
    }\label{2}
\end{figure*}

In this section, the proposed Face R-FCN (See Figure \ref{2}) is described in detail. Since our framework is based on the R-FCN, we refer the reader to \cite{RFCN} for more technical details.

We improve the R-FCN framework for targeting face detection in three aspects.
First, we introduce additional smaller anchors and modify the position sensitive RoI pooling to a smaller size for suiting the detection of the tiny faces. 
Second, we propose to use position-sensitive average pooling instead of normal average pooling for the last feature voting in R-FCN, which leads to an improved embedding. 
Third, multi-scale training strategy and on-line Hard Example Mining (OHEM) strategy \cite{OHEM} are adopted for training. In the testing phase, we also ensemble the multi-scale detection results to improve the performance. 
The details of the proposed approach are described as follows.

\subsection{R-FCN Based Architecture}

R-FCN \cite{RFCN} is a region-based fully convolutional network initially proposed for object detection. Unlike other region-based detectors such as Faster RCNN \cite{FastRCNN}, R-FCN constructs a deeper fully convolutional network without increasing the speed overhead by shared computation on the entire image. R-FCN builds upon 101-layer ResNet \cite{ResNet101}, consists of a region proposal network (RPN) and a R-FCN module in contrast to R-CNN module that presented in Faster R-CNN \cite{FastRCNN}. 

ResNet architecture in R-FCN plays the role of feature extractor. It is common knowledge that ResNet construct a very deep network which is able to extract highly representative image features. These features hold much larger receptive field where tiny face detection can be benefited from the context information. From the feature maps that output by the fundamental ResNet, RPN generates a batch of the region of interests (RoIs) according to the anchors. These RoIs further are fed into two sibling position sensitive RoI pooling layer in R-FCN module to produce class score maps and bounding box prediction maps. In the end of R-FCN, two global average pooling are applied on both class score maps and bounding box prediction maps respectively for aggregating the class scores and bounding box predictions. 

There are two major advantages that we adopt R-FCN over R-CNN. Firstly, the position sensitive RoI pooling ingeniously encodes position information into each RoI by pooling group of feature maps to a certain location of the output score maps; Secondly, without unnaturally injecting fully connected layers into ResNet architecture, the feature maps of R-FCN are trained more expressive and easier for the network to learn the class score and bounding box of faces. 

Based on R-FCN, We make several effective modifications for improving detection performance. For better describing tiny faces, we introduce more anchors with smaller scales (say, from 1 to 64). These smaller anchors are very helpful for sufficiently capturing the extremely tiny faces. Besides, we set smaller pooling size for position sensitive RoI pooling to reduce redundant information, and refine the following voting scheme (average pooling) to be position sensitive average pooling, which will be described in the following section. Finally, we apply atrous convolution in the last stage of ResNet to keep the scale of feature maps without losing the contextual information in larger receptive field\ignore{large receptive field}.

\subsection{Position-Sensitive Average Pooling}

In the original R-FCN work, global average pooling is adopted to aggregate the features after position-sensitive RoI pooling into a single dimension. This operation leads to the uniform\ignore{equal} contribution of each position of the face. However, the contribution of each part of the face may be non-uniformed\ignore{not equal} for detection. For example, in terms of face recognition, eyes usually are paid more attentions than mouth which has been verified by experiments in \cite{ARdatabase}. 
Intuitively, We believe such assumption that distinct regions on the face have different importance should also hold in face detection. Hence, we propose to perform weighted average for each area of the output of position sensitive RoI pooling in order to re-weight the region, which is called position-sensitive average pooling.

Formally, let $\bar X = \{ {X_i}|i = 1,2,...,M\}$ denote the output $M$ feature maps of a position-sensitive RoI pooling layer, and ${X_i} = \{ {x_{i,j}}|j = 1,2,...,N^2 \}$ denote the $i_{th}$ feature map, where $N$ denotes the size of the pooled feature map. Position-sensitive average pooling calculates the weighted average value of the feature responses to get the pooling feature $Y = \{ {y_i}|i = 1,2,...,M\}$ from $\bar X$, where $y_i$ is denoted as:

\begin{equation}
\label{equ:intra}
{y_i} = \frac{1}{{{N^2}}}\sum\limits_{j = 1}^{{N^2}} {w_{j} x_{i,j}},
\end{equation}
where $w_{j}$ denotes the weight for the $j$-th position.
Note that position-sensitive average pooling can be thought as performing feature embedding on every location of responses followed by average pooling. Hence, it is very convenient to implement position-sensitive average pooling on most of the popular deep neural network frameworks.

\subsection{Multi-Scale Training and Testing}	

Inspired by \cite{facercnn}, we perform multi-scale training and testing strategy to improve performance. In the training phase, we resize the shortest side of the input to 1024 or 1200 pixels. This training strategy keeps our model being robust on detecting the target at the different scale, especially on tiny faces. 
On-line Hard Example Mining (OHEM) \cite{OHEM} is a simple yet effective technique for bootstrapping. During training, we also apply OHEM on negative samples and set the positive and negative samples ratio to 1:3 in each mini-batch.
In the testing phase, we build an image pyramid for each test image. Each scale in the pyramid is independently tested. The results from various scales are eventually merged together as the final result of the image.

\section{Experiments}


We perform evaluation on two public-domain face detection benchmarks: the WIDER FACE dataset \cite{wider} and the FDDB dataset \cite{fddb}. 
The WIDER FACE dataset has a total collection of 393,703 labeled face in 32,203 images, of which 40\% are used for training, 10\% for validation and 50\% for testing. Specifically, the validation set and the test set are divided into three subsets (Easy, Medium, and Hard) for evaluation based on different level of difficulties, as defined in \cite{wider}. 
The FDDB dataset contains 5,171 labeled faces in 2,845 images. 
Example images of WIDER FACE and FDDB are shown in Figure \ref{5}.


\subsection{Implementation Details}
Our training hyper-parameters are similar to Face R-CNN \cite{facercnn}. Different from Face R-CNN, we initialize our network with the pre-trained weights of 101-layer ResNet trained on ImageNet. Specifically, we freeze the general kernels (weights of few layers at the beginning) of the pre-trained model throughout the entire training process in order to keep the essential feature extractor trained on ImageNet. 

In terms of the RPN stage, Face R-FCN enumerates multiple configurations of the anchor in order to accurately search for faces. We combine a range of multiple scales and aspect ratios together to construct multi-scale anchors. These anchors then map to the original image to calculate the IoU scores with the ground truth for further picking up with following rules: First, the anchors with highest IoU score are strictly kept as positive; Second, the anchors with IoU score above 0.7 are assigned as positive; Third, If the anchors have IoU score that is lower than 0.3, they are marked as negative. The R-FCN is then trained on the processed anchors (proposals) where the positive samples and negative samples are defined as IoU greater than 0.5 and between 0.1 and 0.5 respectively. The RPN and R-FCN are both learned jointly with the softmax loss and the smooth L1 loss.

Non-maximum suppression (NMS) is adopted for regularizing the anchors with certain IoU scores. The proposals are processed by OHEM to train with hard examples. We set the 256 for the size of RPN mini-batch and 128 for R-FCN respectively. Approximate joint training strategy is applied for training in the end-to-end fashion.

We utilize multi-scale training where the input image is resized with bilinear interpolation to various scales (say, 1024 or 1200). In the testing stage, multi-scale testing is performed by scale image into an image pyramid for better detecting on both tiny and general faces.





\subsection{Comparison on Benchmarks}

\subsubsection{WIDER FACE}

\begin{figure*}
  \centering

  \subfigure[Val: easy]
  {
    \label{fig:2:a}
    \includegraphics[width=6cm, keepaspectratio]{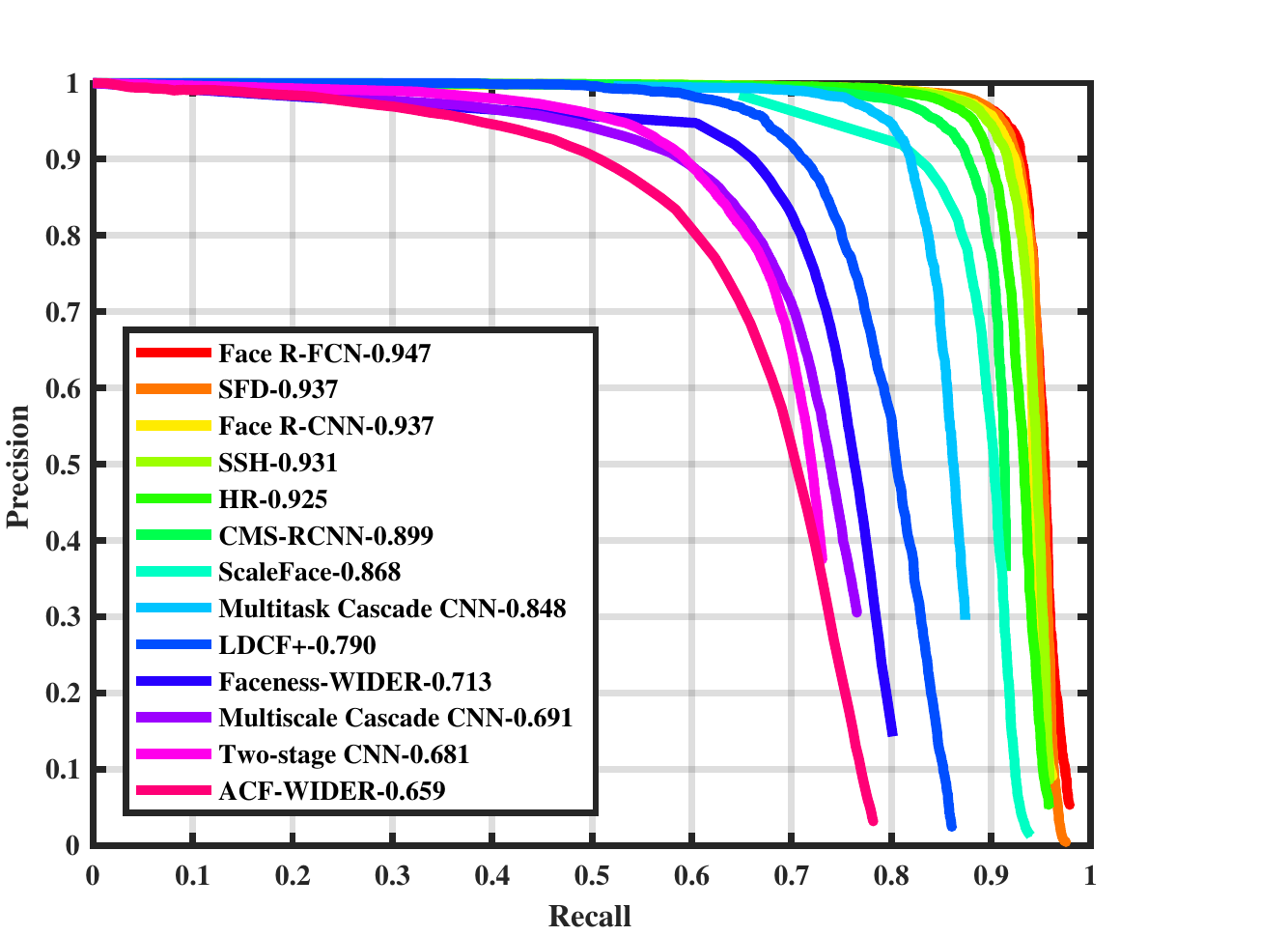}
  }
  \subfigure[Test: easy]	
  {
    \label{fig:2:b}
    \includegraphics[width=6cm, keepaspectratio]{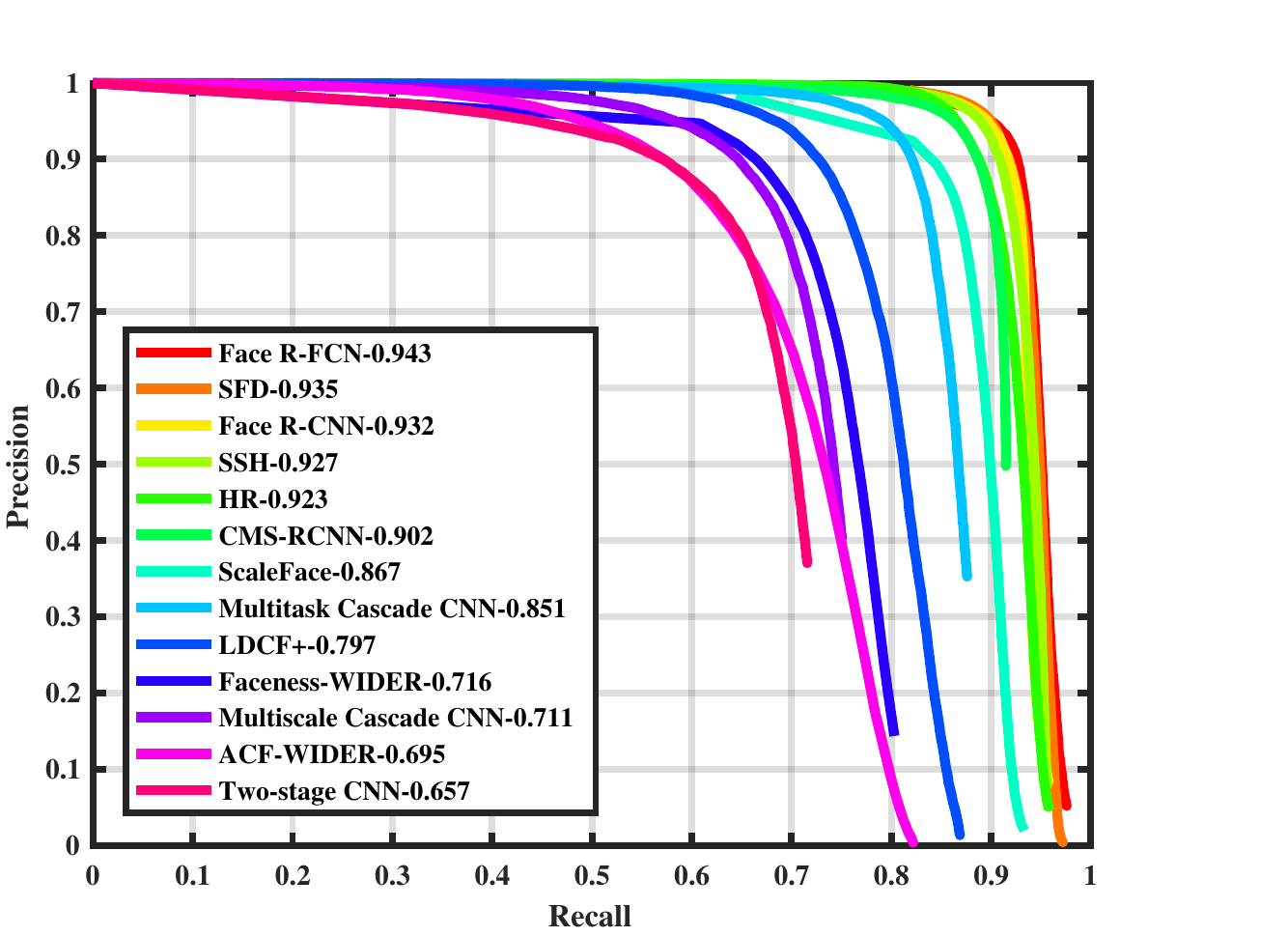}
  }
  \subfigure[Val: medium]
  {
  \label{fig:2:c}
  \includegraphics[width=6cm, keepaspectratio]{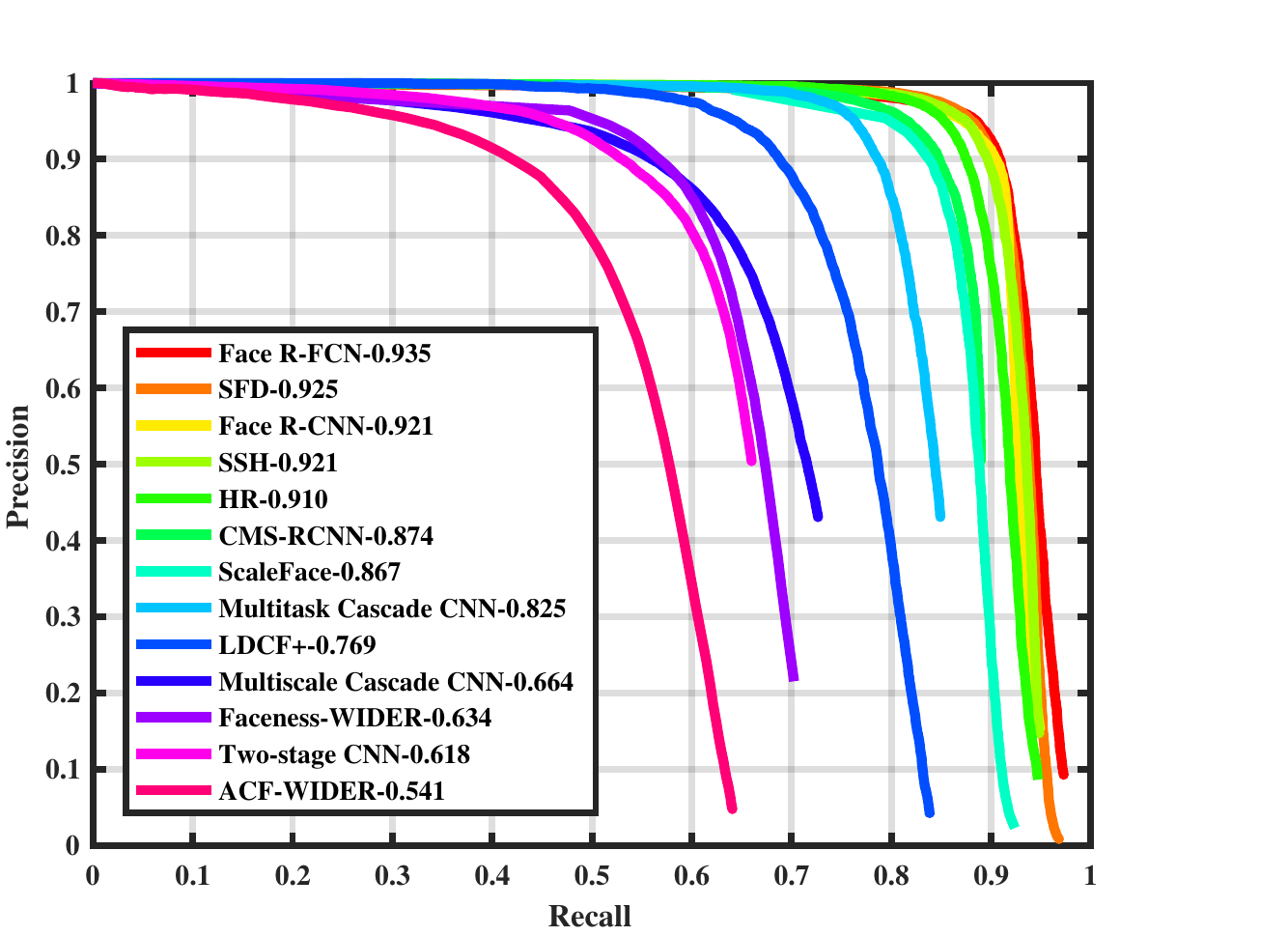}
  }
  \subfigure[Test: medium]	
  {
  \label{fig:2:d}
  \includegraphics[width=6cm, keepaspectratio]{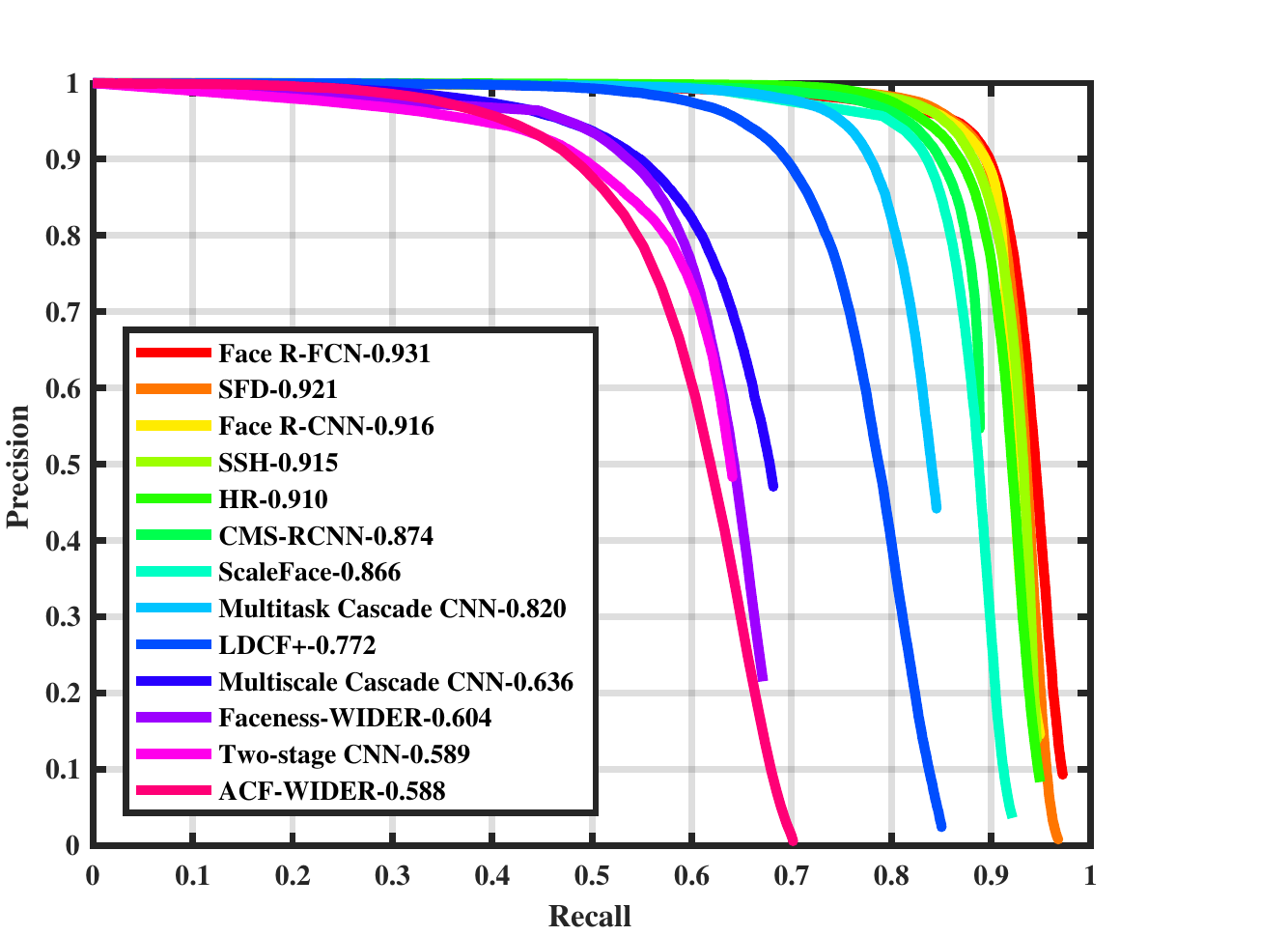}
  }
  \subfigure[Val: hard]
  {
  \label{fig:2:e}
  \includegraphics[width=6cm, keepaspectratio]{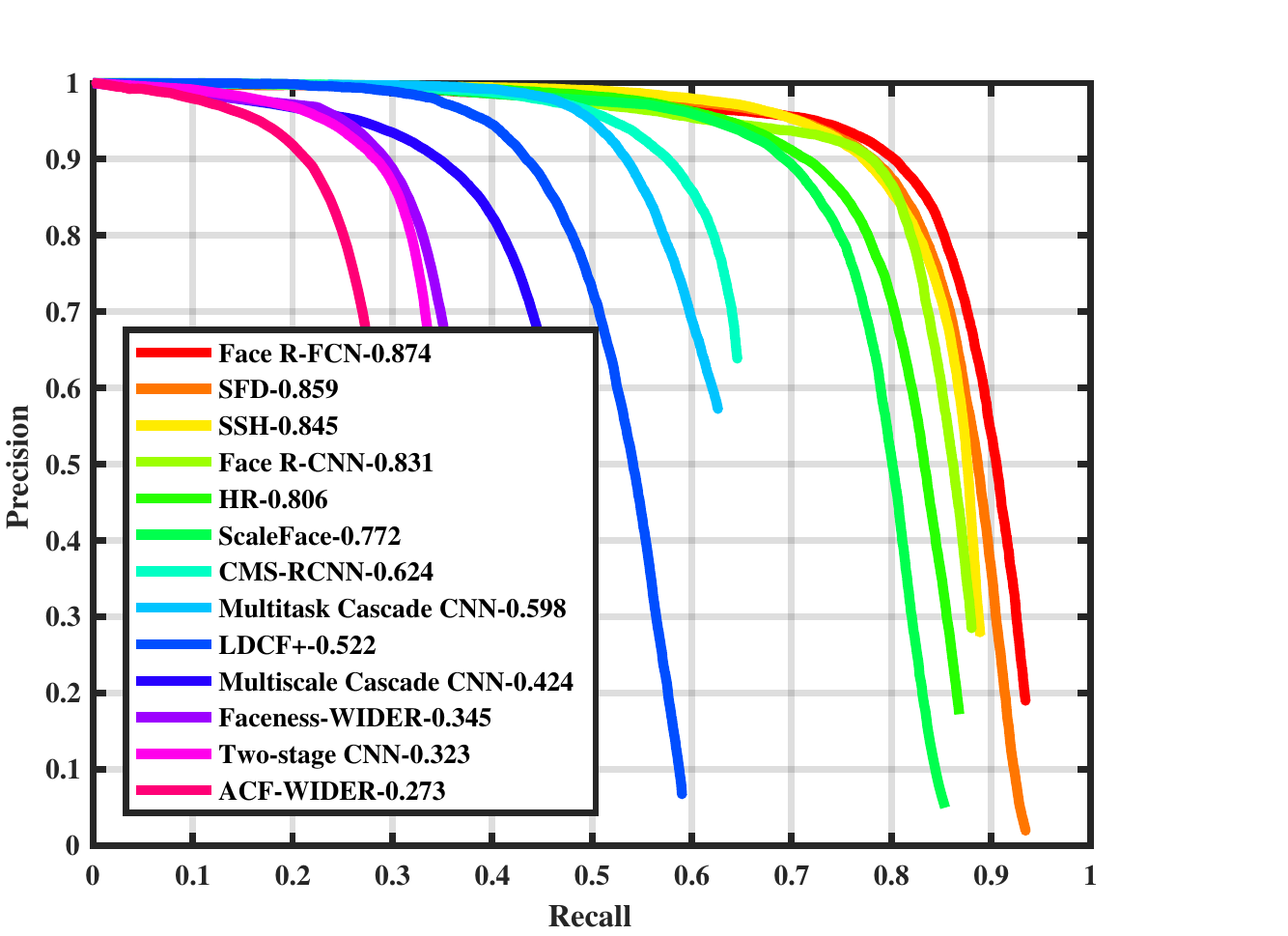}
  }
  \subfigure[Test: hard]	
  {
  \label{fig:2:f}
  \includegraphics[width=6cm, keepaspectratio]{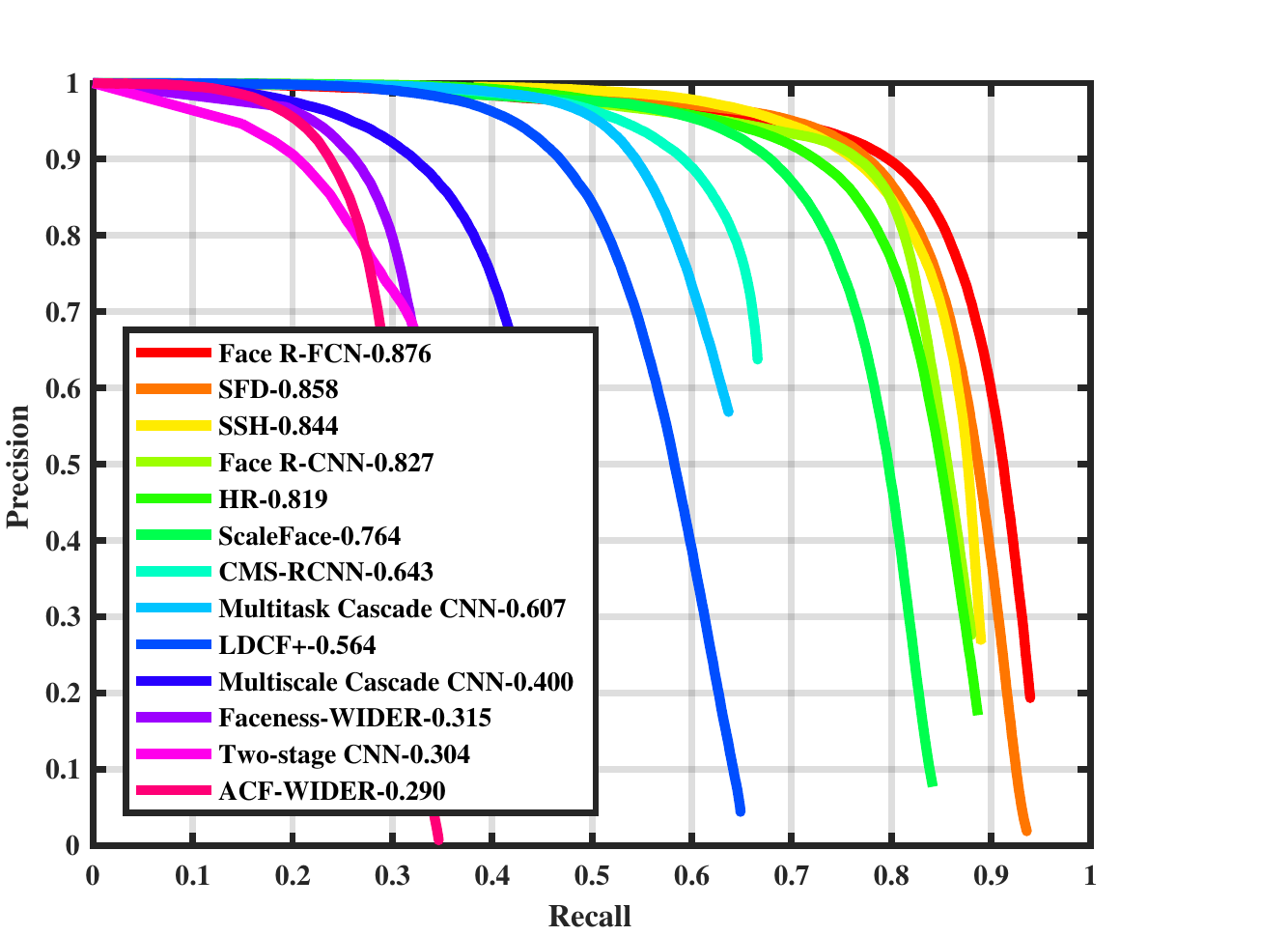}
  }
  \caption{Precision-Recall curves on WIDER FACE's validation set and test set. All of these methods use the same Scenario-Int criterion \cite{wider}. Face R-FCN shows the superior performance over the prior methods across the three subsets (easy, medium and hard) in both validation and test sets. Best viewed in color.
}\label{3}

\end{figure*}

We train our model on the training set of WIDER FACE and perform
evaluation on the validation set and test set following the Scenario-Int criterion \cite{wider}.
As illustrated in Figure \ref{3}, our proposed approach consistently wins the 1st place across the three subsets on both the validation set and test set of WIDER FACE and significantly outperforms the existing results \cite{ssh,sfd,HR,cmsrcnn,spl,vj2,faceness,wider}.
In particular, on WIDER FACE hard subset, our approach is superior to the prior best-performing one \cite{sfd} by a clear margin, which demonstrates the robustness of our algorithm.

\subsubsection{FDDB}


There are two evaluation protocols for evaluating the FDDB dataset: one is 10-fold cross-validation and the other is unrestricted training (using the data outside FDDB for training). Our experiments strictly follow the protocol for unrestricted training.

\begin{figure*}
  \centering

  \subfigure[]
  {
    \label{fig:3:a}
    \includegraphics[width=12cm, keepaspectratio]{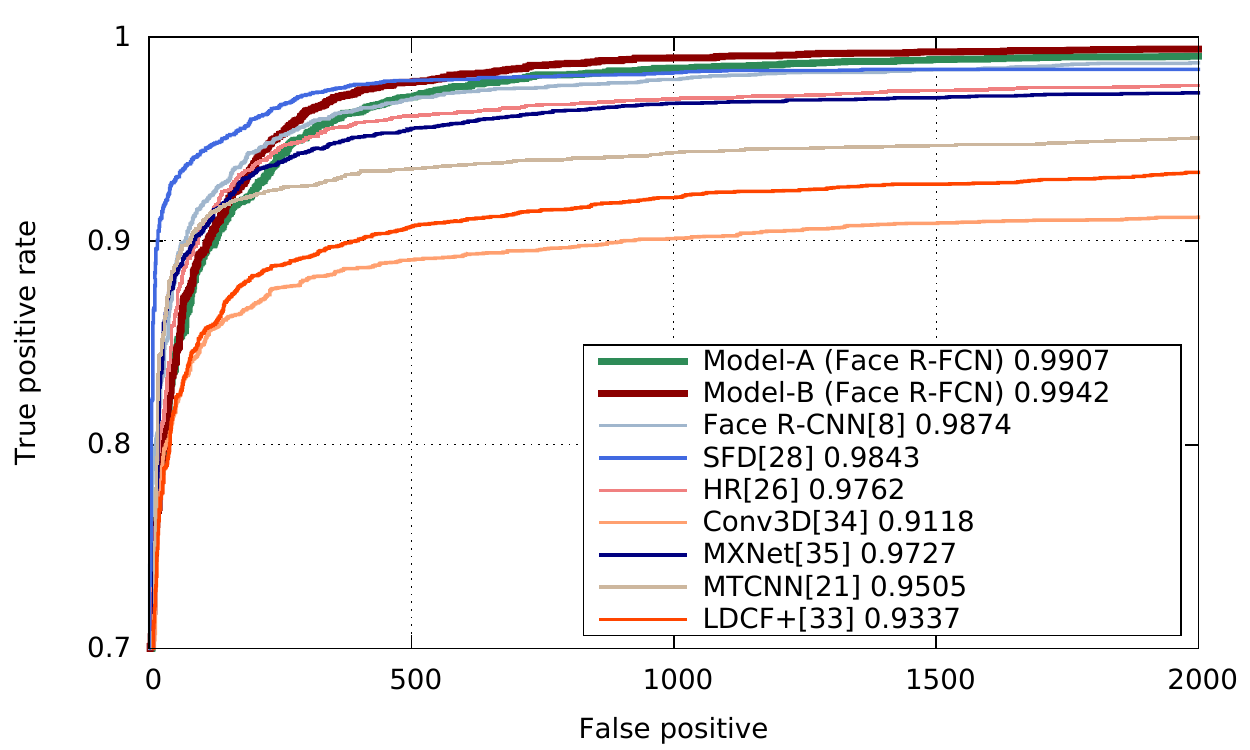}
  }
  \subfigure[]
  {
    \label{fig:3:a}
    \includegraphics[width=12cm, keepaspectratio]{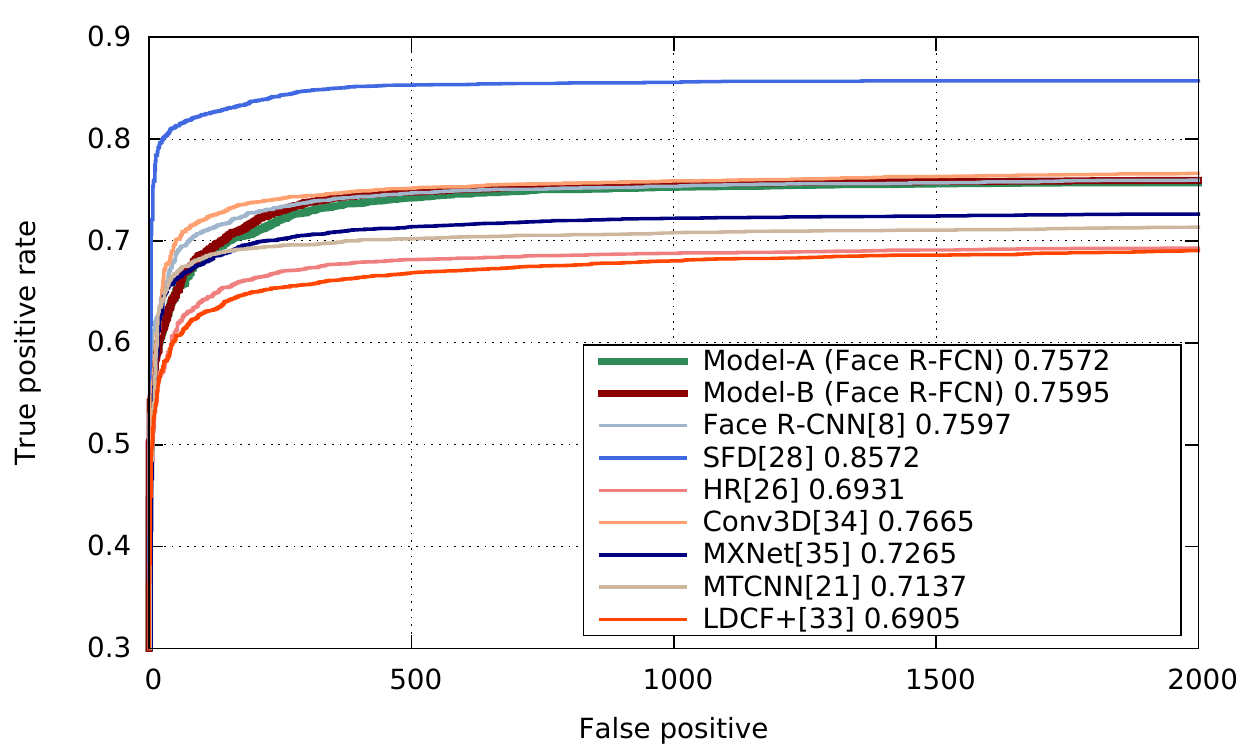}
  }

  \caption{Evaluation of our results on the FDDB published methods. We show the ROC curves on the (a) Discrete ROC curve and (b) Continuous ROC curve.
Model-A and Model-B is trained by WIDER FACE's training set and a augmented private dataset respectively. We show the true positive rate at 2000 false positives for each model. Best viewed in color.}\label{4}
\end{figure*}

\begin{figure*}
  \centering

  \subfigure[]
  {
    \label{fig:5:a}
	\includegraphics[width=11cm, keepaspectratio]{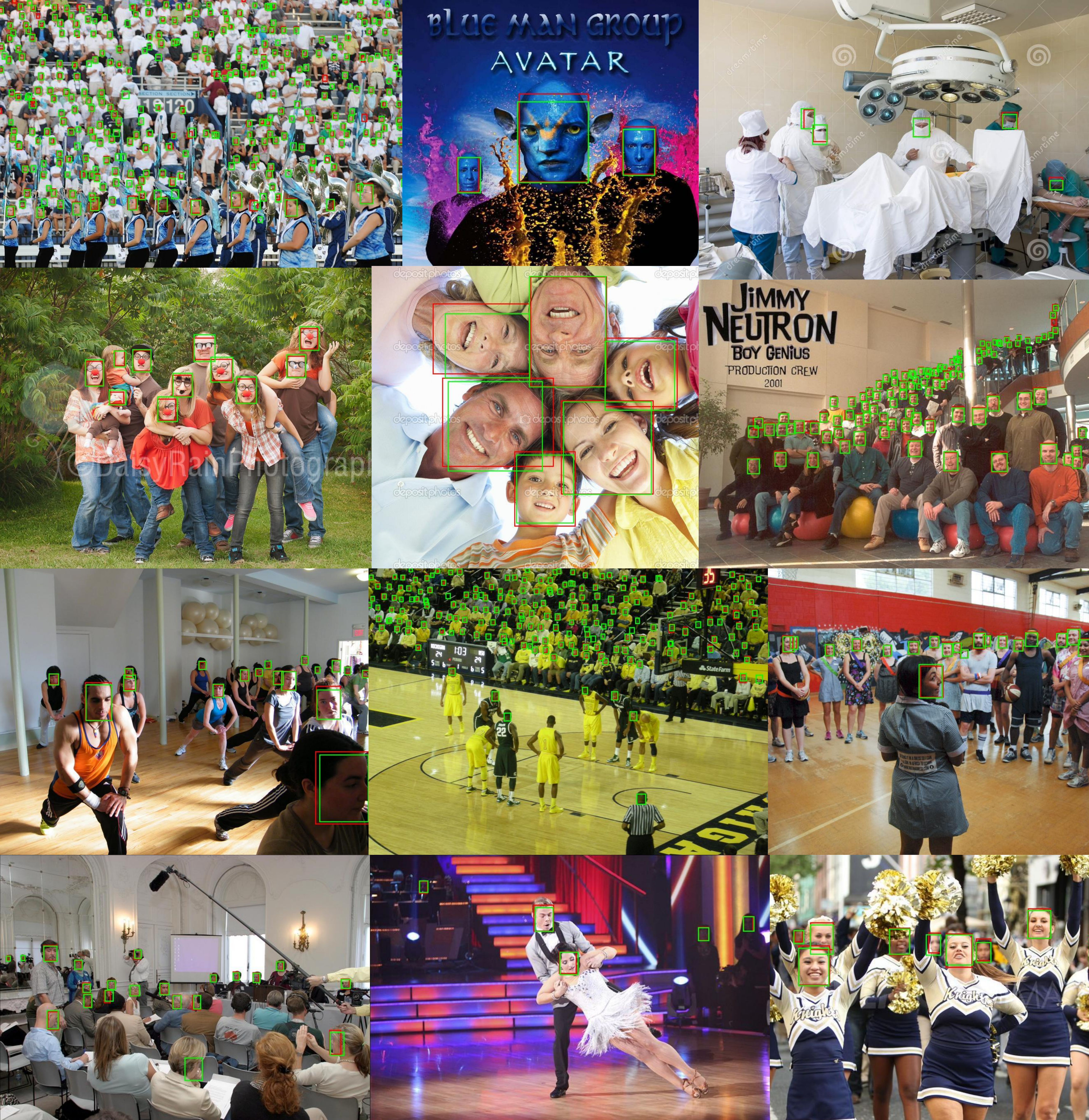}
  }
  \subfigure[]
  {
    \label{fig:5:b}
	\includegraphics[width=10cm, keepaspectratio]{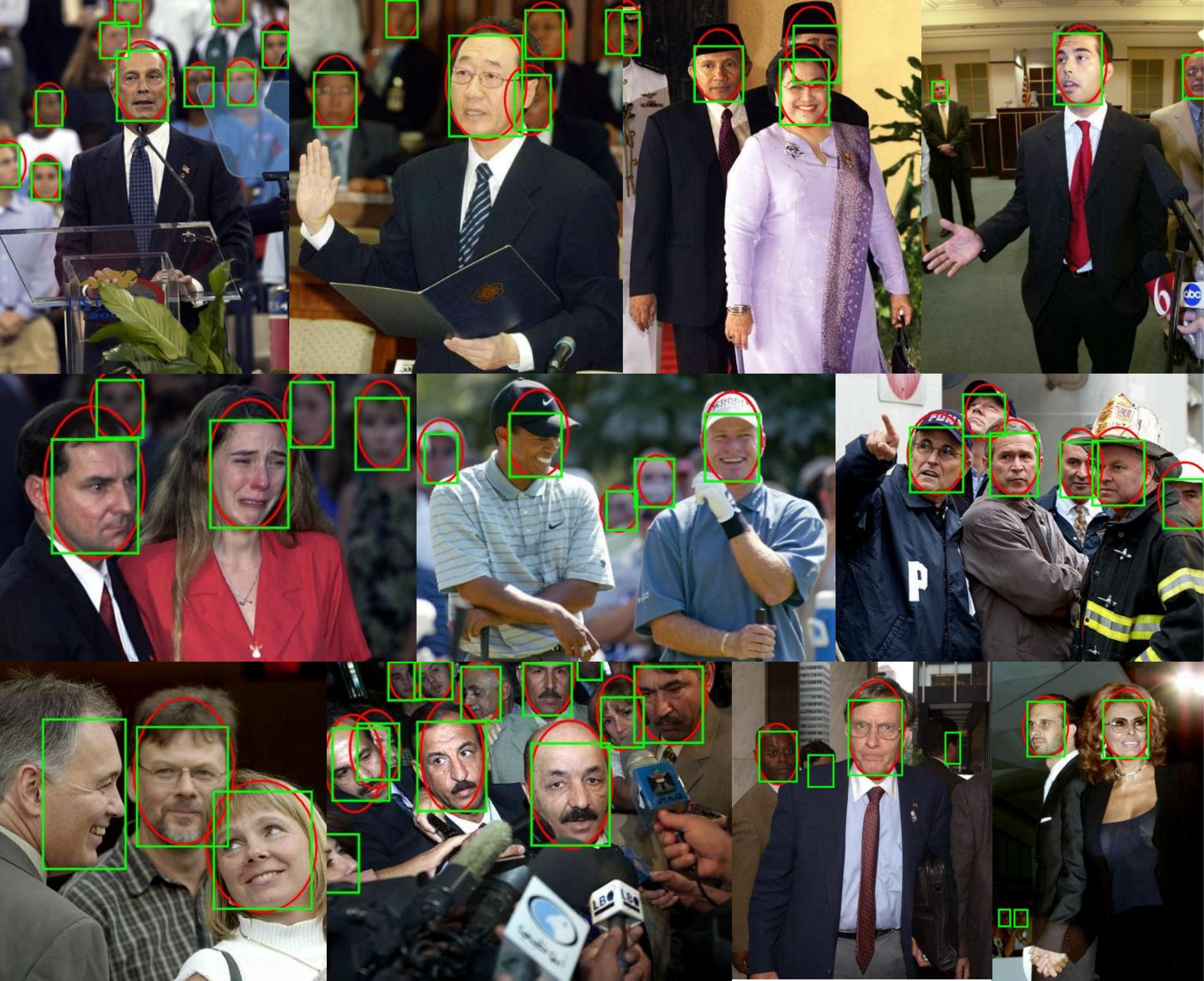}
  }

  \caption{
	Examples of our detected results on the (a) WIDER FACE validation set and (b) FDDB. The green frames in the image represent the face detection results while the red frames or ellipses represent the ground-truth annotations.
Note that in the last row of (b), some of human faces detected by Face R-FCN have not been annotated as ground truth.
    }\label{5}
\end{figure*}

We use the training set of the WIDER FACE dataset to train our model (denoted as Model-A in Figure \ref{4}) and compare against the recently published top approaches \cite{sfd,HR,spl,ldcf,conv3d,mxnet} on FDDB. All of these approaches use the protocol for unrestricted training defined in \cite{fddb}. The discrete ROC curves and continuous ROC curves of these approaches are plotted in Figure \ref{4}. From Figure \ref{4}, it is clearly that Face R-FCN consistently achieves the impressive\ignore{state-of-the-art} performance in terms of both the discrete ROC curve and continuous ROC curve. Our discrete ROC curve is superior to the prior best-performing method\ignore{one} \cite{sfd,facercnn}. We also obtain the best true positive rate of the discrete ROC curve at 1000/2000 false positives (98.49\%/99.07\%). For the reason that we do not optimize our method to regress the elliptical ground truth in FDDB dataset, our continuous ROC curve is lower than the first place \cite{sfd} and slightly lower than \cite{facercnn,conv3d}. Additionally, one of the factors that may affect the performance of Face R-FCN demonstrated in the last row of \ref{fig:5:b}: the false positive bounding boxes in images exactly contain faces from human perspective where these faces have not been annotated as ground truth. This factor partly leads to the lower performance comparing with \cite{facercnn,conv3d,sfd}. But the competitive result we achieved is still noticeable.  

Furthermore, We expand the training dataset by augmenting with a privately collected dataset and use the enlarged dataset to train a more discriminative face detector (denoted as Model-B). The discrete and continuous ROC curves of Model-B are also plotted in Figure \ref{4}. 
As expected, the performance of Face R-FCN is further improved. Finally, we obtain the true positive rate 98.99\% of the discrete ROC curve at 1000 false positives and 99.42\% at 2000 false positives, which are new state-of-the-art among all the published methods on FDDB.


\section{Conclusion}

Face detection is a fundamental problem in vision task.  
In this technical report, we propose a powerful face detection approach named Face R-FCN by integrating R-FCN and several sophisticated techniques for better detecting faces and boosting overall performance. By reasoning the drawbacks of R-CNN and R-FCN, we explore the details and invent new designs to improve the popular detection framework specifically for face detection. The proposed approach is evaluated on the challenging WIDER FACE dataset and FDDB dataset. Our experimental results demonstrate the superiority of our approach over the state-of-the-arts. These innovations are inspired from past experience and we expect our innovations will be easy to generally applied to the future face detection architectures as past experience.

{\small
\bibliographystyle{unsrt}
\bibliography{refbib}
}

\end{document}